\RequirePackage{fix-cm}
\documentclass[smallextended]{svjour3}       
\smartqed  
\usepackage{algorithm}
\usepackage{algpseudocode}
\usepackage{graphicx}
\usepackage{amssymb}
\usepackage{algorithm}
\usepackage{algpseudocode}
\usepackage{amsmath}
\usepackage{multirow}
\usepackage{hhline}
\usepackage{color}
\newcommand\tab[1][.2cm]{\hspace*{#1}}

\begin{document}

\title{Semi-Supervised Learning for Detecting Human Trafficking
}

\titlerunning{Semi-Supervised Learning for Detecting Human Trafficking}        

\author{Hamidreza Alvari $^{1}$         \and
	Paulo Shakarian $^{1}$ \and
	J.E. Kelly Snyder $^{2}$ 
}

\authorrunning{Alvari et al.} 

\institute{$^{1}$ Arizona State University\\
	Tempe, Arizona \at
	\email{\{halvari,shak\}@asu.edu}           
	\and
	$^{2}$ Find Me Group\\
	Tempe, Arizona \at
	\email{kelly@findmegroup.org}
}

\date{Received: date / Accepted: date}

\maketitle

\begin{abstract}
Human trafficking is one of the most atrocious crimes and among the challenging problems facing law enforcement which demands attention of global magnitude. In this study, we leverage textual data from the website ``Backpage"-- used for classified advertisement-- to discern potential patterns of human trafficking activities which manifest online and identify advertisements of high interest to law enforcement. Due to the lack of ground truth, we rely on a human analyst from law enforcement, for hand-labeling a small portion of the crawled data. We extend the existing Laplacian SVM and present $S^3VM-R$, by adding a regularization term to exploit exogenous information embedded in our feature space in favor of the task at hand. We train the proposed method using labeled and unlabeled data and evaluate it on a fraction of the unlabeled data, herein referred to as \textit{unseen} data, with our expert's further verification. Results from comparisons between our method and other semi-supervised and supervised approaches on the labeled data demonstrate that our learner is effective in identifying advertisements of high interest to law enforcement.  

\keywords{Human Trafficking \and Backpage \and Semi-Supervised Support Vector Machines \and Laplacian Support Vector Machines}
\end{abstract}

\section{Introduction}
\label{intro}
According to the United Nation~\cite{unodc}, human trafficking is defined as the modern slavery or the trade of humans mostly for the purpose of sexual exploitation and forced labor, via different improper ways including force, fraud and deception. The United States' Trafficking Victim Protection Act of 2000 (TVPA 2000)~\cite{tvpa2000} was the first U.S. legislation passed against human trafficking. Human trafficking has ever since received increased national and societal concern~\cite{HT_report2015} but still demands persistent fight against from all over the globe. No country is immune and the problem is rapidly growing with little to no law enforcement addressing the issue. This problem is amongst the challenging ones facing law enforcement as it is difficult to identify victims and counter traffickers.

Before the advent of the Internet, human traffickers were under risks of being arrested by law enforcement while advertising their victims on streets~\cite{desplaces92}. However, move to the Internet has made it easier and less dangerous for sex sellers~\cite{nicholas2012} as they no longer needed to advertise on the streets. There are now a plethora of websites that host and provide sexual services under categories of escort, adult entertainment, massage services, etc., which help sex sellers and buyers maintain their anonymity. Although some services such as the Craiglist's adult section and myredbook.com were shut down recently, there are still many websites such as the Backpage.com that provide such services and many new are frequently created. Traffickers even use dating and social networking websites such as the Twitter, Facebook, Instagram and Tinder to reach out to sex buyers and their followers. Although the Internet has presented new trafficking related challenges for law enforcement, it has also provided readily and publicly available rich source of information which could be gleaned from online sex advertisements for fighting this crime~\cite{kennedy2012predictive}. However, the problem is we lack the ground truth and obtaining the labels through hand-labeling is indeed tedious and expensive even for a small subset of data-- this is the point where the semi-supervised setting comes in handy. 

Despite considerable attention which has been devoted to studying supervised, unsupervised and semi-supervised learning settings via different applications~\cite{mitchell2006learning,beigi2016signed,backstrom2011supervised,alvari2016identifying,beigi2016exploiting,mitchell1997machine,beigi2014leveraging}, semi-supervised learning, i.e., learning from labeled and unlabeled examples, is still one of the most interesting yet challenging problems in the machine learning community~\cite{belkin2006manifold}. The idea is simple though-- we shall have an approach that makes a better use of unlabeled data to boost performance. This is pretty close to the most natural learning that occurs in the world. For the most part, we as humans are exposed only to a small number of labeled instances; yet we successfully generalize well by effective utilization of a large amount of unlabeled data. This motivates us to use unlabeled samples to improve recognition performance while developing classifiers.

In this article, expanding on our previous work~\cite{alvari2016non}, we use the data crawled from the adult entertainment section of the website Backpage.com and extend the existing Laplacian SVM framework~\cite{belkin2006manifold} to detect escort advertisements of high interest to law enforcement. Here, we merely focus on the online advertisements although the Internet has triggered many other activities including attracting the victims, communicating with customers and rating the escort services. We thus highlight several contributions of the current research as follows.

\begin{enumerate} 
	\item Based on the literature, we created different groups of features that capture the characteristics of potential human trafficking activities. The less likely human trafficking related posts were then filtered out using these features. We also conducted a feature importance analysis to demonstrate how these features contribute to the proposed learner.
	\item We extended the Laplacian SVM~\cite{belkin2006manifold} and proposed the semi-supervised support vector machine learning algorithm, $S^3VM-R$. In particular, we incorporated additional information of our feature space as a regularization term into the standard optimization formulation with regard to the Laplacian SVM. We also used geometry of the underlying data as an intrinsic regularization term in Laplacian SVM.
	\item We trained our model on both of the \textit{labeled} and \textit{unlabeled} data and sent back the identified human trafficking related advertisements to an expert from law enforcement for further verification. We then validated our approach on a small subset of the unlabeled data (i.e. \textit{unseen} data) with further verification of the expert.
	\item We performed comparisons between our approach and several semi-supervised and supervised baselines on both of the labeled and unseen data (so-called \textit{blind} evaluation).
	\item We demonstrated the effect of varying different hyperparameters used in our learner on its performance.
\end{enumerate}

The rest of the paper is organized as follows. In Section 2, we review the prior studies on human trafficking. Section 3 covers our data preparation, feature engineering,  unsupervised filtering and expert assisted labeling. We detail our semi-supervised learning approach in Section 4 by deriving the required equations. Section 5 provides in-depth explanation of our experiments. Section 6 concludes the paper by providing future research directions.

\section{Related Work}
Recently, several studies have examined the role of the Internet and related technology in facilitating human trafficking~\cite{hughes2005demand,hughes2002use,latonero2011human}. For example, the work of~\cite{hughes2005demand} studied how closely sex trafficking is intertwined with new technologies. According to~\cite{hughes2002use}, sexual exploitation of women and children is a global human right crisis that is being escalated by the use of new technologies. Researchers have studied relationships between new technologies and human trafficking and advantages of the Internet for sex traffickers. For instance, findings from a group of experts from the Council of Europe demonstrated that the Internet and sex industry are closely interlinked and volume and content of the material on the Internet promoting human trafficking are unprecedented~\cite{latonero2011human}.

One of the earliest works which leveraged data mining techniques for online human trafficking was~\cite{latonero2011human}, wherein the authors conducted data analysis on the adult section of the website Backpage.com. Their findings confirmed that female escort post frequency would increase in Dallas, Texas, leading up to the Super Bowl 2011 event. In a similar attempt, other studies~\cite{dominique2015,2016arXiv160205048M} have investigated impact of large public events such as the Super Bowl on sex trafficking by exploring advertisement volume, trends and movement of advertisements along with the scope and volume of demand associated with such events. The work of~\cite{dominique2015} for instance, concluded that large events such as the Super Bowl which attract significant amount of concentration of people in a relatively short period of time and in a confined urban area, could be a desirable location for sex traffickers to bring their victims for commercial sexual exploitation. Similarly, the data-driven approach of~\cite{2016arXiv160205048M} showed that in some but not all events, one can see a correlation between occurrence of the event and statistically significant evidence of an influx of sex trafficking activity. Also, certain studies~\cite{conf/semweb/SzekelyKSPSYKNM15} have tried to build large distributed systems to store and process available online human trafficking data in order to perform entity resolution and create ontological relations between entities. 

Beyond these works, the work of~\cite{2015arXiv150906659N} studied the problem of isolating sources of human trafficking from online advertisements with a pairwise entity resolution approach. Specifically, they used phone number as a strong feature and trained a classifier to predict if two ads are from the same source. This classifier was then used to perform entity resolution using a heuristically learned value for the score of classifier. Another work of~\cite{kennedy2012predictive} used Backpage.com data and extracted most likely human trafficking spatio-temporal patterns with the help of law enforcement. Note that unlike our method, this work did not employ any machine learning methodologies for automatically identifying human trafficking related advertisements. The work of~\cite{doi:10.1080/23322705.2015.1015342} also deployed machine learning for the advertisement classification problem, by training a supervised learning classifier on labeled data (based on phone numbers of known traffickers) provided by a victim advocacy group. We note that while phone numbers can provide a very precise set of positive labeled data, there are clearly many posts with previously unseen phone numbers. 

In contrast, we do not solely rely on phone numbers for labeling our data. Instead, our expert analyze each post's content to identify whether it is human trafficking related or not. To do so, we first filter out most likely advertisements using several feature groups and pass a small sample to the expert for hand-labeling. Then, we train our semi-supervised learner on both of the labeled and unlabeled data which in turn lets us evaluate our approach on new coming (unseen) data later. We note that our semi-supervised approach can also be used as a complementary method to procedures such as those described in~\cite{doi:10.1080/23322705.2015.1015342} as we can significantly expand the training set for use with supervised learning. 

Finally, note that our current research is different from our previous work~\cite{alvari2016non} and we list the key nuances here:
\begin{itemize}
	\item In this study we experiment with a much larger dataset. To obtain such dataset, we use the same raw data from~\cite{alvari2016non}, but this time with slight modifications of the thresholds that were used for filtering out less likely human trafficking related advertisements.
	\item As opposed to our previous research which deployed only one feature space, in this work, two feature spaces that have complementary roles to each other are used. 
	\item In this paper we present a new framework based on the existing Laplacian SVM~\cite{belkin2006manifold}, by adding a regularization term to the standard optimization problem and solving the new optimization equation derived from there. In contrast,~\cite{alvari2016non} utilized the off-the-shelf graph based semi-supervised learner, LabelSpreading method~\cite{Zhou04learningwith}, without any further manipulation of the original approach.
	\item Unlike~\cite{alvari2016non} in which we did not compare our method with other approaches, this work compares our proposed framework against other semi-supervised and supervised learners. Also unlike our previous work in which only \textit{one} group of human trafficking related advertisements were passed to \textit{two} experts for validation, here in order to reduce the inconsistency, \textit{two} control groups of advertisements--those of interest to law enforcement and those of not-- are sent to only \textit{one} expert for verification. 	
\end{itemize} 

\section{Data Preparation}
\begin{figure}[!ht]
	\caption{A real post from Backpage.com. To ensure anonymity, the personal information has been intentionally obfuscated.}
	\centering
	\includegraphics[width=0.7\textwidth]{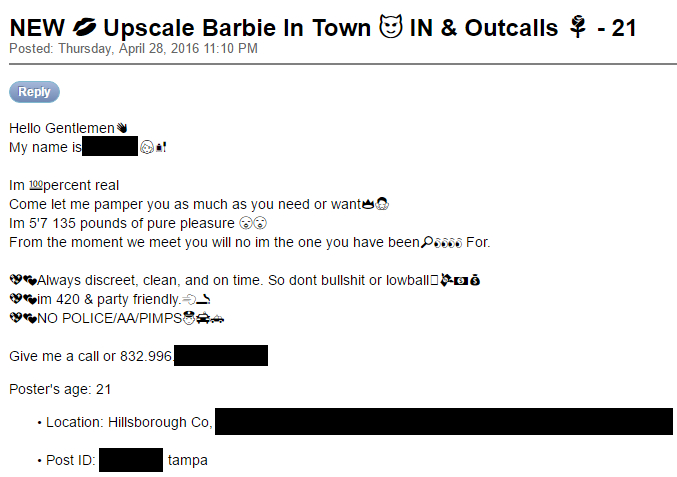}
	\label{fig:post1}
\end{figure}

\begin{figure}[!ht]
	\caption{Choropleth map of phone frequency with respect to the different states. Darker colors show higher frequencies.}
	\centering
	\includegraphics[width=0.8\textwidth]{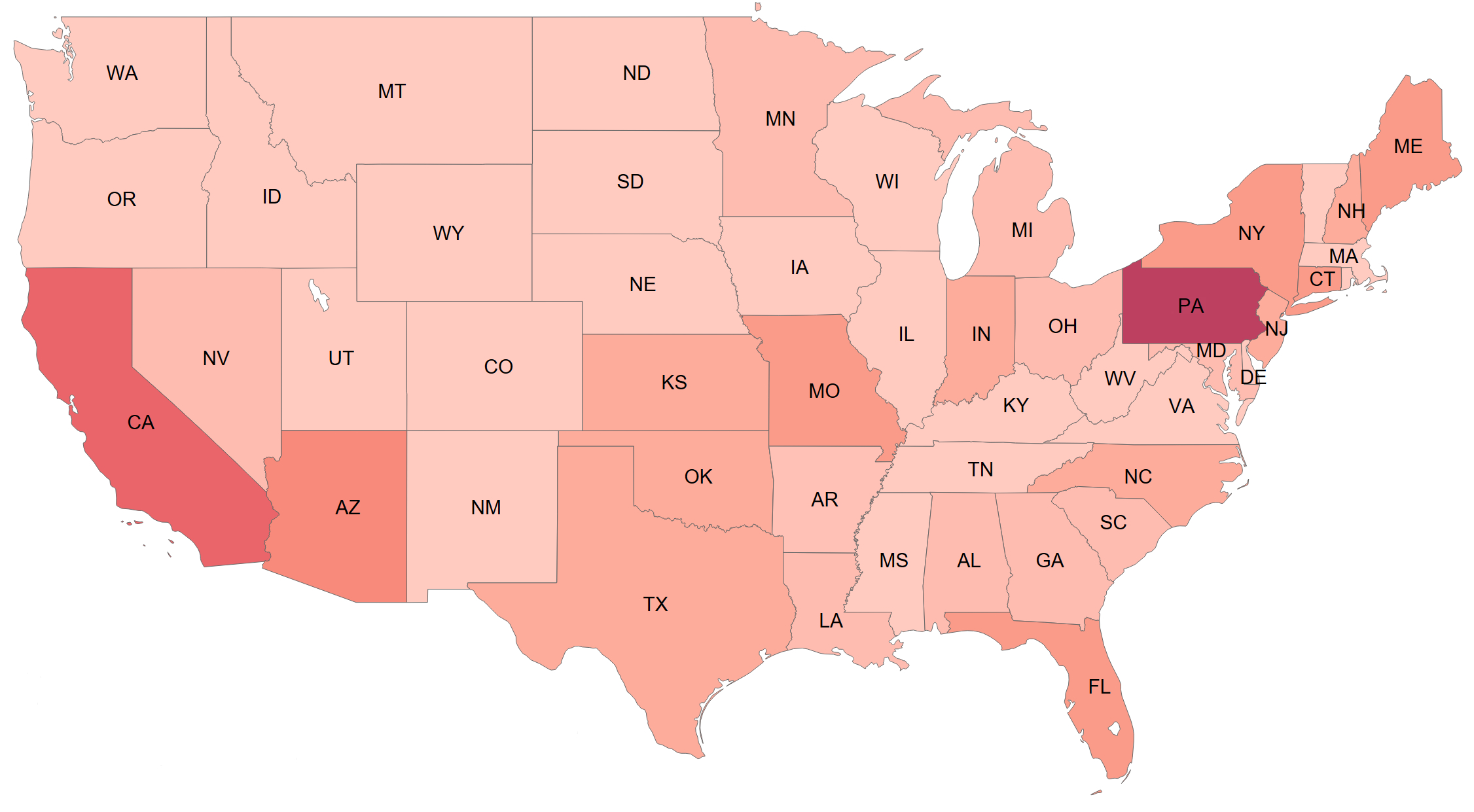}
	\label{fig:dist}
\end{figure}

We collected about 20K publicly available listings from the U.S. posted on Backpage.com in March, 2016. Each post includes a title, description, time stamp, poster's age, poster's ID, location, image, and sometimes video and audio. The description usually lists the attributes of the individual(s) and contact phone numbers. In this work, we only focus on the textual component of the data. This free-text data required significant cleaning due to a variety of issues common to textual analytics (i.e. misspellings, format of phone numbers, etc.). We also acknowledge that the information in data could be intentionally inaccurate, such as poster's name, age and even physical appearance (e.g. bra cup size, weight). Figure~\ref{fig:post1} shows an actual post from Backpage.com. To illustrate geographic diversity of the listings, we use the Tableau{\footnote{https://www.tableau.com/}} software to visualize choropleth map of phone frequency with respect to the different states in Figure~\ref{fig:dist}, wherein darker colors mean higher frequencies.

Next, we will explain most important characteristics of potential human trafficking advertisements which are captured by our feature groups. 

\begin{figure}[!ht]
	\caption{An evidence of human trafficking. The boxes and numbers in red, indicate the features and their corresponding group numbers (see also Table~\ref{tb:feat}).}
	\centering
	\includegraphics[width=0.7\textwidth]{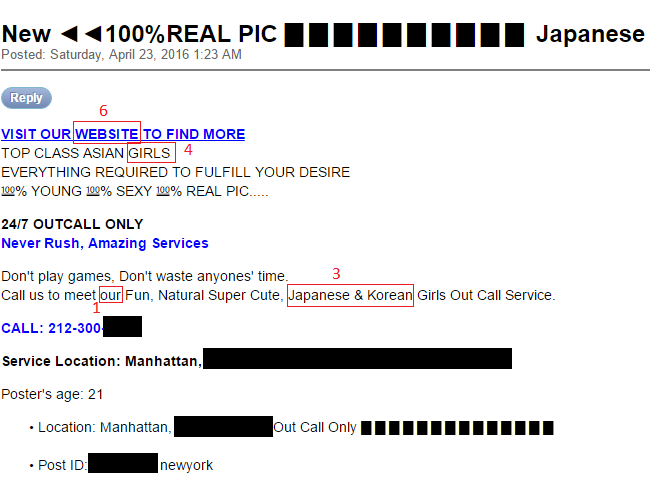}
	\label{fig:post2}
\end{figure}

\subsection{Feature Engineering}
Though many advertisements on Backpage.com are posted by posters selling their own services without coercion and intervention of traffickers, some do exhibit many common trafficking triggers. For example, in contrast to Figure~\ref{fig:post1}, Figure~\ref{fig:post2} shows an advertisement that could be an evidence of human trafficking. This advertisement indicates several potential properties of human trafficking, including advertising for multiple escorts with the first individual coming from Asia and very young. In what follows, such common properties of human trafficking related advertisements are discussed in more detail.

Inspired by the literature, we define and extract 6 groups of features from advertisements  (see Table~\ref{tb:feat}). These features could be amongst the strong indicators of human trafficking. Let us now briefly describe each group of features used in our work. Note each feature listed here is ultimately treated as a \textit{binary} variable.

\begin{table}[t]
	\centering
	\caption{Different features and their corresponding groups.}
	\begin{tabular}{|c|l|c|}
		\hline
		\textbf{No.} & \textbf{Feature Group} & \textbf{Ref.}\\ \hline\hline
		\multirow{5}{*}{1} & \textbf{Advertisement Language Pattern} & \cite{kennedy2012predictive,Li:2008:IKC:1478784,conf/setn/KanarisKS06}\\
		& \tab - Third person language & \\
		& \tab - First person plural pronouns & \\
		& \tab - Kolmogorov complexity & \\ 
		& \tab - \textit{n}-grams (1) & \\
		& \tab - \textit{n}-grams (2) & \\
		& \tab - \textit{n}-grams (3) & \\ 
		& & \\
		2 & \textbf{Words and Phrases of Interest} & \cite{hetter2012,Lloyd2012,Goodman2011} \\
		& & \\
		3 & \textbf{Countries of Interest} & \cite{HT_report2015}\\
		& & \\
		4 & \textbf{Multiple Victims Advertised} & \cite{kennedy2012predictive}\\
		& & \\
		5 & \textbf{Victim Weight} & \cite{tvpa2000,weightchart}\\
		& & \\
		\multirow{2}{*}{6} & \textbf{Reference to Website or Spa Massage Therapy} & \cite{kennedy2012predictive}\\
		& \tab - Reference to a website & \\
		& \tab - Reference to a Spa Massage Therapy & \\ \hline
	\end{tabular}
	\label{tb:feat}
\end{table}

\subsubsection{Advertisement Language Pattern} The first group consists of different language related features. For the first and second features, we identify posts which have third person language (more likely to be written by someone other than the escort) and posts which contain first person plural pronouns such as `we' and `our' (more likely to be an organization)~\cite{kennedy2012predictive}.

To ensure their anonymity, traffickers would deploy techniques to generate diverse information and hence make their posts look more complicated. They usually do this to avoid being identified by either human analysts or automated programs. Thus, to obtain the third feature we take an approach from complexity theory, namely \textit{Kolmogorov complexity}, which is defined as length of shortest program to reproduce a string of characters on a universal machine such as the Turing Machine~\cite{Li:2008:IKC:1478784}. Since the Kolmogorov complexity is not computable, we approximate the complexity of an advertisement content by first removing stop words and then computing entropy of the content~\cite{Li:2008:IKC:1478784}. To illustrate this, let $X$ denote the content and $x_i$ be a given word in the content. We use the following equation~\cite{shannon2001mathematical} to calculate the entropy of the content and thus approximate the Kolmogorov complexity of $X$:
\begin{equation}
K(X) \approx -\sum_{i=1}^n{P(x_i)\log_2 P(x_i)}
\end{equation}

We expect higher values of the entropy correspond to human trafficking. Finally, we discretize the result by using the threshold of 4 which was found empirically in our experiments.

For the next features, we use word-level \textit{n}-grams to find common language patterns of advertisements. This particular choice is because of the fact that character-level \textit{n}-grams have already shown to be useful in detecting unwanted content for spam detection~\cite{conf/setn/KanarisKS06}. We set $n=4$ and use the range of (4,4) to compute normalized \textit{n}-grams (using TF-IDF) of each advertisement content. We ultimately create a matrix whose rows and columns correspond to the advertisements contents and their associated 4-grams, respectively. We rank all elements of this matrix in a descending order and pick the top 3 ones. Finally for each advertisement content, 3 elements with the column numbers associated with the top elements are chosen. This way, 3 more features will be added to our feature set. Overall, we have 6 features related to the language of the advertisement.

\subsubsection{Words and Phrases of Interest} Despite the fact that advertisements on Backpage.com do not directly mention sex with children, customers who prefer children know to look for words and phrases such as ``\textit{sweet}, \textit{candy}, \textit{fresh}, \textit{new in town}, \textit{new to the game}"~\cite{hetter2012,Lloyd2012,Goodman2011}. We thus investigate within the posts to see if they contain such words as they could be highly related with human trafficking in general.

\subsubsection{Countries of Interest} We identify if the individual being escorted is coming from other countries such as those in Southeast Asia (especially from China, Vietnam, Korea and Thailand, as we observed in our data)~\cite{HT_report2015}.

\subsubsection{Multiple Victims Advertised} Some advertisements advertise for multiple women at the same time. We consider the presence of more than one victim as a potential evidence of organized human trafficking~\cite{kennedy2012predictive}.

\subsubsection{Victim Weight} We take into account the weight of the individual being escorted as a feature (if it is available). This information is particularly useful assuming that for the most part, lower body weights (under 115 lbs) correlate with smaller and underage girls~\cite{tvpa2000,weightchart} and thereby human trafficking.

\subsubsection{Reference to Website or Spa Massage Therapy} The presence of a link in the advertisement either referencing to an outside website (especially infamous ones) or spa massage therapy could be an indicator of more elaborate organization~\cite{kennedy2012predictive}. In particular, in case of spa therapy, we observed many advertisements interrelated with advertising for young Asian girls and their erotic massage abilities. Therefore, the last group of features has two binary features for presence of any website and spa.

Finally, in order to extract all of the above features, we first clean the original data and conduct preprocessing. By applying these features, we draw a random sample of 3,543 instances out of our dataset for further analysis to see if they are evidences of human trafficking-- this is described in the next section. 

\subsection{Unsupervised Filtering}
\begin{figure}[!ht]
	\caption{2-D projection of the entire set of the filtered data.}
	\centering
	\includegraphics[width=0.6\textwidth]{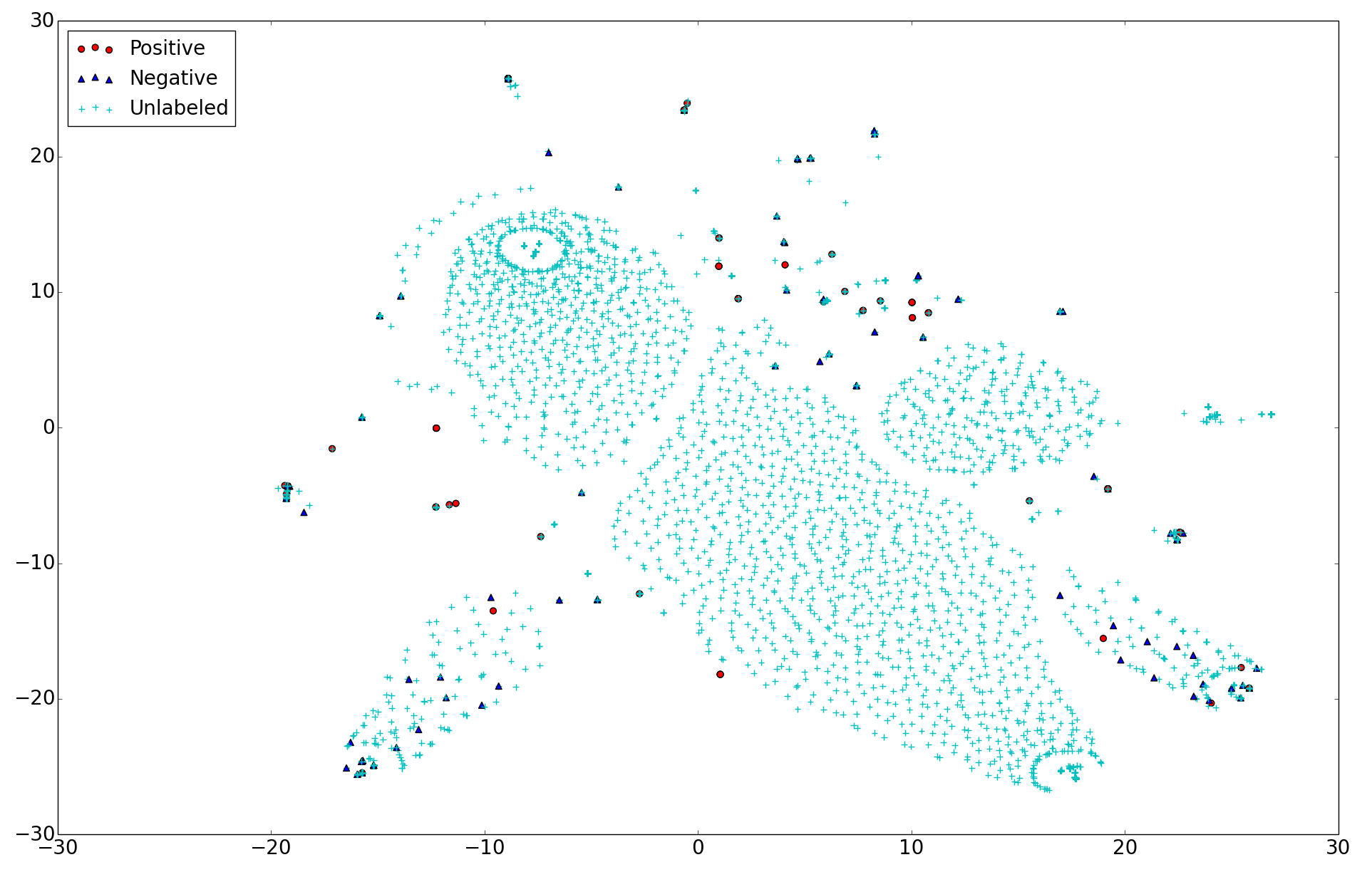}
	\label{fig:all}
\end{figure}

Having detailed our feature set, we now construct a feature vector for each instance by creating a vector of 12 binary features that correspond to the important characteristics of human trafficking. Hereafter, we refer to this feature space, as our \textit{first} feature space and denote it with $\mathcal{F}_1$. As mentioned earlier, we draw 3,543 instances from our raw data by filtering out those that do not posses any of the binary features. We will refer to this as our \textit{filtered} dataset. For the sake of visualization, a 2-D projection (using the t-SNE transformation~\cite{maaten2008visualizing}) of the filtered dataset is depicted in Figure~\ref{fig:all}. The purpose of this figure is to demonstrate how hard it is for basic clustering techniques such as the K-means, to correctly assign labels to unlabeled instances using only few existing labeled ones.

Now, we shall define our \textit{second} feature space, namely $\mathcal{F}_2$, which will be used to compute geometry of the underlying data. Note that our proposed framework will utilize both of the feature spaces in the form of regularization terms, to detect advertisements of high interest to law enforcement. After conducting standard preprocessing techniques on the filtered dataset, we build $\mathcal{F}_2$ by transforming the filtered data into a 3,543$\times$3,543 matrix of TF-IDF similarity features. Each entry in this matrix simply shows the similarity between a pair of advertisements in our filtered dataset.

Note that since we lack the ground truth, we would rely on a human analyst (expert) for labeling the listings as either `of interest' or `of not interest' to law enforcement. In the next section, we select a smaller yet finer grain subset of this data to be sent to the expert. This alleviates the burden of the tedious work of hand-labeling.

\subsection{Expert Assisted Labeling} 
We first obtain a sample of 200 listings from the filtered dataset. This set of listings was labeled by our expert from law enforcement who is specialized in this type of crime. From this subset, the law enforcement professional identified 70 instances to be of interest to law enforcement and the rest to be not human trafficking related. However, we are still left with a large amount of the unlabeled examples (3,343 instances) in our dataset. The ratio of the labeled to unlabeled instances in our dataset is very small (about 0.06). The statistics of our dataset is summarized in Table~\ref{tb:st}.
\begin{table}[t]
	\centering
	\caption{Description of the dataset.}
	\begin{tabular}{|l|c|c|}
		\cline{1-3}
		\textbf{Name}          & \multicolumn{2}{c|}{\textbf{Value}}\\
		\hhline{===}
		Raw      & \multicolumn{2}{c|}{20,822}  \\ \cline{1-3}
		Filtered & \multicolumn{2}{c|}{3,543} \\ \cline{1-3}
		Unlabeled & \multicolumn{2}{c|}{3,343} \\ \cline{1-3}
		Labeled  & \textbf{Positive} & \textbf{Negative} \\ \cline{2-3}
		 & 70  & 130 \\
		 \cline{1-3}
	\end{tabular}
	\label{tb:st}
\end{table}

\section{Semi-Supervised Learning Framework}
Here, we first introduce some preliminary notations necessary for the rest of the discussion and then outline our proposed semi-supervised approach, $ُS^3VM-R$, for detecting online human trafficking. Note as said earlier, our framework is an extension to the existing Laplacian SVM~\cite{belkin2006manifold}. In particular, we incorporated another regularization term into the standard Laplacian SVM to leverage the additional information of our first feature space and then solved the associated optimization problem.  Consequently, similar notation is adopted throughout the following section. Furthermore, we shall once again note that our current research does not utilize any off-the-shelf graph based semi-supervised leaner in contrast to our previous research~\cite{alvari2016non}. 

\subsection{Technical Preliminaries}
We assume a set of $l$ labeled pairs $\{(x_i,y_i)\}_{i=1}^l$ and an unlabeled set of $u$ instances $\{x_{l+i}\}_{i=1}^u$, where $x_i\in\mathbb{R}^n$ and $y_i\in\{+1,-1\}$. Recall for the standard soft-margin support vector machine, the following optimization problem is solved:
\begin{equation}
	\min_{f_\theta\in \mathcal{H}_k} \gamma||f_\theta||_k^2 + C_l \sum_{i=1}^{l}H_1(y_if_\theta(x_i))
\end{equation} 

In the above equation, $f_\theta(\cdot)$ is a decision function of the form $f_\theta(\cdot)=w.\mathbf{\Phi}(\cdot)+b$ where $\theta=(w,b)$ are the parameters of the model, and $\mathbf{\Phi(\cdot)}$ is the feature map which is usually implemented using the kernel trick~\cite{cortes1995support}. Also, the function $H_1(\cdot)=\max(0,1-\cdot)$ is the Hinge Loss function. 

The classical Representer theorem~\cite{belkin2005manifold} suggests that solution to the optimization problem exists in a Hilbert space $\mathcal{H}_k$ and is of the following form:

\begin{equation}
f_\theta^*(x) = \sum_{i=1}^{l}\alpha_i^*\mathbf{K}(x,x_i)
\end{equation}

\noindent where $\mathbf{K}$ is the $l\times l$ Gram matrix over labeled samples. Equivalently, the above problem can be written as:

\begin{align}
&\min_{w,b,\epsilon} \frac{1}{2}||w||_2^2 + C_l \sum_{i=1}^{l}\epsilon_i 
\end{align}

\begin{align}
&~s.t.~~~y_i(w.\mathbf{\Phi}(x_i)+b)\geq 1-\epsilon_i, ~ i=1,...,l \nonumber\\
&~~~~~~~~\epsilon_i\geq 0, ~ i=1,...,l
\end{align}

Next, we will use the above optimization equation as our basis to derive the formulations for our proposed semi-supervised learner.

\subsection{The proposed Method}
The basic assumption behind semi-supervised learning methods is to leverage unlabeled instances in order to restructure hypotheses during the learning process. In this paper, exogenous information extracted from both of our feature spaces is further exploited to make a better use of the unlabeled examples. To do so, we first introduce matrix $\mathbf{F}$ in $\mathcal{F}_1$ and over both of the labeled and unlabeled samples with $\mathbf{F}_{ij}$ defined as follows:

\begin{equation}
\mathbf{F}_{ij}=\frac{1}{n_f}(\mathbf{\Phi}(x_i)\cdot\mathbf{\Phi}(x_j))
\end{equation}

\noindent where $n_f$ is the number of features in $\mathcal{F}_1$ (here, $n_f=12$). We force the instances $x_i$ and $x_j$ in our dataset to have same label if they both possess same features. To account for this, a regularization term is added to the standard equation and the following optimization is solved: 

\begin{equation}
\min_{f_\theta \in \mathcal{H}_k} \frac{1}{2}\sum_{i=1}^{l}\textbf{F}_{ij}||f_\theta(x_i)-f_\theta(x_j)||_2^2 = \mathbf{f}^T_\theta\mathcal{L}^T\mathbf{f}_\theta
\end{equation} 

\noindent where $\mathbf{f}=[f(x_1), ..., f(x_{l+u})]^T$ and $\mathcal{L}$ is the Laplacian matrix based on $\mathbf{F}$ given by $\mathcal{L}=\mathbf{D}-\mathbf{F}$, and $\mathbf{D}_{ii}=\sum_{j=1}^{l+u}\mathbf{F}_{ij}$. The intuition here is that any two instances which are composed of same features are more likely to have same labels than others. Next, by solving a similar optimization problem, we are able to capture data geometry in $\mathcal{F}_2$ as $\mathbf{f}^T_\theta\mathcal{L}'^T\mathbf{f}_\theta$ (also referred to as the intrinsic smoothness penalty term~\cite{belkin2006manifold}). Here, $\mathcal{L}'$ is the Laplacian of matrix $\mathbf{A}$ associated with the data adjacency graph $\mathbf{G}$ in $\mathcal{F}_2$.


We construct $\mathbf{G}$ with $(l+u)$ nodes in $\mathcal{F}_2$, and by adding an edge between each pair of nodes $\langle i,j \rangle$, if the edge weight $W_{ij}$ exceeds a given threshold. For computing the edge weights, we use the heat kernel~\cite{grigor2006heat} as a function of the Euclidean distance between two samples in $\mathcal{F}_2$, hence we set $W_{ij}=\exp^{-||x_i-x_j||^2/4t}$.

Following the notations used in~\cite{belkin2006manifold} and by including our regularization term as well as the intrinsic smoothness penalty term, we would extend the standard equation by solving the following optimization:

\begin{equation}\label{eq:opt1}
\min_{f_\theta\in \mathcal{H}_k} \gamma||f_\theta||_k^2 
+
C_l \sum_{i=1}^{l}H_1(y_if_\theta(x_i)) 
+ 
C_r\textbf{f}_\theta^T\mathcal{L}\textbf{f}_\theta
+  C_s \textbf{f}_\theta^T\mathcal{L'}\textbf{f}_\theta
\end{equation}

Note one typical value for the smoothness penalty coefficient $C_s$ is $\frac{\gamma_I}{(l+u)^2}$, where $\frac{1}{(l+u)^2}$ is a natural scale factor for empirical estimate of the Laplace operator and $\gamma_I$ is a regularization term~\cite{belkin2006manifold}. Again, solution in $\mathcal{H}_k$ would be in the following form:

\begin{equation}
f_\theta^*(x) = \sum_{i=1}^{l+u}\alpha_i^*\mathbf{K}(x,x_i)
\end{equation}

Here $\mathbf{K}$ is the $(l+u)\times(l+u)$ Gram matrix over all samples. The equation~\ref{eq:opt1} could be then written as follows:

\begin{align}
&\min_{\alpha,b,\epsilon} \frac{1}{2}\alpha^T\mathbf{K}\alpha + C_l \sum_{i=1}^{l}\epsilon_i + \frac{C_r}{2}\alpha^T\mathbf{K}\mathcal{L}\mathbf{K}\alpha + \frac{\gamma_I}{2(l+u)^2}\alpha^T\mathbf{K}\mathcal{L}'\mathbf{K}\alpha
\end{align}

\begin{align}
&~s.t.~~~y_i(\sum_{j=1}^{l+u}\alpha_j\mathbf{K}(x_i,x_j)+b)\geq 1-\epsilon_i, ~ i=1,...,l 
\nonumber\\
&~~~~~~~~\epsilon_i\geq 0, ~ i=1,...,l
\end{align}

With introduction of the Lagrangian multipliers $\beta$ and $\gamma$, we write the Lagrangian function of the above equation as follows:

\begin{align}
L(\alpha,\epsilon,b,\beta,\gamma)=\frac{1}{2}\alpha^T\mathbf{K}(I+C_r\mathcal{L}+\frac{\gamma_I}{(l+u)^2}\mathcal{L}')\alpha
+C_l\sum_{i=1}^{l}\epsilon_i
\nonumber\\
-\sum_{i=1}^{l}\beta_i(y_i(\sum_{j=1}^{l+u}\alpha_j\mathbf{K}(x_i,x_j)+b)-1+\epsilon_i) - \sum_{i=1}^{l}\gamma_i\epsilon_i
\end{align} 

Obtaining the dual representation, requires taking the following steps:

\begin{align}
\frac{\partial L}{\partial b} = 0 \rightarrow \sum_{i=1}^{l}\beta_iy_i = 0
\end{align}

\begin{align}
\frac{\partial L}{\partial \epsilon_i} = 0 \rightarrow C_l - \beta_i - \gamma_i = 0 \rightarrow 0\leq\beta_i\leq C_l
\end{align}

With the above equations, we formulate the reduced Lagrangian as a function of only $\alpha$ and $\beta$ as follows:

\begin{align}
L^R(\alpha,\beta)=\frac{1}{2}\alpha^T\mathbf{K}(I+C_r\mathcal{L}+\frac{\gamma_I}{(l+u)^2}\mathcal{L}')\alpha
\nonumber\\
-\sum_{i=1}^{l}\beta_i(y_i(\sum_{j=1}^{l+u}\alpha_j\mathbf{K}(x_i,x_j)+b)-1+\epsilon_i)\nonumber\\
\end{align} 

This equation is further simplified as follows:

\begin{align}
L^R(\alpha,\beta)= \frac{1}{2}\alpha^T\mathbf{K}(I+C_r\mathcal{L}+\frac{\gamma_I}{(l+u)^2}\mathcal{L}')\alpha
\nonumber\\
-\alpha^T\mathbf{K}\mathbf{J}^T\mathbf{Y}\beta+\sum_{i=1}^{l}\beta_i
\end{align} 

In the above equation, $\mathbf{J}=[\mathbf{I}~\mathbf{0}]$ is a $l\times(l+u)$ matrix, $\mathbf{I}$ is the $l\times l$ identity matrix and $\mathbf{Y}$ is a diagonal matrix consisting of the labels of the labeled examples.

In the followings, we first take the derivative of $L^R$ with respect to $\alpha$ and then set $\frac{\partial L^R(\alpha,\beta)}{\partial \alpha} = 0$:

\begin{align}
 \mathbf{K}(I+C_r\mathcal{L}+\frac{\gamma_I}{(l+u)^2}\mathcal{L}')\alpha
-\mathbf{K}\mathbf{J}^T\mathbf{Y}\beta = 0 
\end{align}

Accordingly, we obtain $\alpha^*$ by solving the following equation: 

\begin{align}\label{eq:alpha_star}
 \alpha^* = (I+C_r\mathcal{L}+\frac{\gamma_I}{(l+u)^2}\mathcal{L}')^{-1}\mathbf{J}^T\mathbf{Y}\beta^*
\end{align}

Next, we obtain the dual problem in the form of a quadratic programming problem by substituting $\alpha$ back in the reduced Lagrangian function:

\begin{align}\label{eq:beta_star}
\beta^* = {\operatorname{argmax}}_{\beta \in \mathbb{R}^l}~-\frac{1}{2}\beta^T\mathbf{Q}\beta + \sum_{i=1}^{l}\beta_i 
\end{align} 

\begin{align}
 s.t.~~~~
\sum_{i=1}^{l}\beta_iy_i = 0 \nonumber\\
0\leq \beta_i \leq C_l
\end{align}  

\noindent where $\beta = [\beta_1,...,\beta_l]^T \in \mathbb{R}^l$ are the Lagrangian multipliers and $\mathbf{Q}$ is obtained as follows:

\begin{equation}\label{eq:Q}
\mathbf{Q} = \mathbf{YJK}(I+(C_r\mathcal{L}+\frac{\gamma_I}{(l+u)^2}\mathcal{L}')\mathbf{K})^{-1}\mathbf{J}^T\mathbf{Y}
\end{equation}

We summarize the proposed semi-supervised framework in Algorithm 1. Our optimization problem is very similar to the standard optimization problem solved for SVMs, hence we use a standard optimizer for SVMs to solve our problem. 

\begin{algorithm}
	\caption{\textbf{The Proposed Semi-Supervised Framework }}\label{alg:alg1}
	
	\begin{algorithmic}[1]
		\Require $\{(x_i,y_i)\}_{i=1}^l$, $\{x_{l+i}\}_{i=1}^u$, $\mathcal{F}_1$, $\mathcal{F}_2$, $C_l$, $C_r$, $C_s$.
		\Ensure Estimated function $f_\theta: \mathbb{R}^n\rightarrow\mathbb{R}$
		\State \text{Construct matrix $\mathbf{F}$ based on the features in $\mathcal{F}_1$}
		\State \text {Compute the corresponding Laplacian matrix $\mathcal{L}$.}
		\State \text{Construct $\mathbf{A}$ according to the features in $\mathcal{F}_2$.} 
		\State \text{Compute the graph Laplacian matrix $\mathcal{L}'$.}
		\State \text{Construct the gram matrix over all examples using $\mathbf{K}_{ij}=k(x_i,x_j)$} where $k$ is a kernel function.
		\State Compute $\alpha^*$ and $\beta^*$ using Eq.~\ref{eq:alpha_star} and Eq.~\ref{eq:beta_star} and a standard QP solvers.
		\State Compute function $f_\theta^*(x) = \sum_{i=1}^{l+u}\alpha_i^*\mathbf{K}(x,x_i)$
	\end{algorithmic}
\end{algorithm}

\section{Experimental Study}
In this section, we provide a comprehensive analysis of the proposed framework by designing a series of experiments on the filtered dataset. First, we explain several approaches used in this study. Next, various results are discussed: (1) comparisons on the \textit{labeled} data were made between our method and other approaches, (2) experiments were performed on a fraction of the unlabeled data (i.e., unseen data), and the results were further verified by our expert to see what fraction is of interest to law enforcement, (3) \textit{blind} evaluation was conducted to examine other approaches on the unseen data, and finally, (4) experiments were designed to analyze effect of varying different hyperparameters on our method as well as impact of different groups of features in $\mathcal{F}_1$ on our approach.

\subsection{Approaches}

We present results for the following methods:

\begin{itemize}
	\item \textbf{Semi-Supervised}: $S^3VM-R$, Laplacian support vector machines~\cite{belkin2006manifold}, graph inference based label spreading approach~\cite{Zhou04learningwith} with radial basis function (RBF) and K-nearest neighbors (KNN) kernels, and co-training learner~\cite{blum1998combining} with two support vector machines classifiers (SVM).
	\item \textbf{Supervised}: SVM, KNN, Gaussian na\"{i}ve Bayes, logistic regression, adaboost and random forest.
\end{itemize}

For the sake of fair comparison, all algorithms were implemented and run in Python. More specifically, the Python package CVXOPT\footnote{http://cvxopt.org/} was used to implement $S^3VM-R$ and Laplacian support vector machines, and all other approaches were implemented with the help of the Scikit-learn\footnote{http://scikit-learn.org/stable/} package in Python. Note for those methods that require special tuning of parameters, we performed grid search to choose the best set of parameters. Before going any further, we first define main parameters used in each method and then demonstrate their best values picked by our grid search. The discussion on the effect of varying the hyperparameters on our learner is provided in the section~\ref{sec:hypr}.

\begin{itemize}
	\item $S^3VM-R$: we set the penalty parameter as $C_l=0.6$ and the regularization parameters $C_r=0.2$ and $C_s=0.2$. Linear kernel was used in our approach.
	\item \textit{Laplacian SVM}: we used linear kernel and set the parameters $C_l=0.6$ and $C_s=0.6$.
	\item \textit{LabelSpreading (RBF)}: RBF Kernel was used and $\gamma$ was set to the default value of 20.  
	\item \textit{LabelSpreading (KNN)}: KNN kernel was used and the number of neighbors was set to 5.
	\item \textit{Co-training (SVM)}: we followed the algorithm introduced in~\cite{blum1998combining} and used two SVM as our classifiers. For both SVMs we set the tolerance for stopping criteria to 0.001 and the penalty parameter $C=1$.
	\item \textit{SVM}: tolerance for stopping criteria was set to the default value of 0.001. Penalty parameter $C$ was set to 1 and linear kernel was used.
	\item \textit{KNN}: number of neighbors was set to 5.
	\item \textit{Gaussian NB}: there were no specific parameter to tune.
	\item \textit{Logistic regression}: we used the `\textit{l}2' penalty. We also set the parameter $C=1$ (the inverse of regularization strength) and tolerance for stopping criteria to 0.01.
	\item \textit{Adaboost}: number of estimators was set to 200 and we also set the learning rate to 0.01.
	\item \textit{Random forest}: we used 200 estimators and the `entropy' criterion was used.	
\end{itemize}

\subsection{Classification Results}
Here, we first evaluate the entire set of approaches on a small portion of the data for which we already know the labels, i.e., the \textit{labeled} examples. We note that expert-generated judgmental labeling might be error-prone, though it is served as a surrogate to the ground truth problem.  

We used 10-fold cross-validation on the labeled data in the following way. We first divided the set of the labeled samples into 10 different sets of approximately equal size. Each time we held one set out for validation (by removing their labels and adding them to the unlabeled samples) and used the remaining along with the unlabeled samples for the training--this was performed for all approaches for the sake of fair comparison. Finally, we reported the average of 10 different runs, using different combinations of the feature spaces and various evaluation metrics, including the area under curve (AUC), accuracy, precision, recall and F1-score. In table~\ref{tb:auc_acc}, we reported the average AUC and accuracy for each method and each feature space. On the other hand, for precision, recall and F1-score, we reported separate results for each feature space, in tables~\ref{tb:prf_F1}-\ref{tb:prf_F1_F2}, respectively. Note, each of these tables includes separate scores for the positive and negative classes. In general, we observed the followings: 

\begin{itemize}
	\item Overall, our approach achieved highest performance on $\mathcal{F}_1$ (tables~\ref{tb:auc_acc} and~\ref{tb:prf_F1}) and $\{\mathcal{F}_1, \mathcal{F}_2\}$ (table~\ref{tb:prf_F1_F2}), in terms of all metrics. However it did not perform well using solely $\mathcal{F}_2$ (table~\ref{tb:prf_F2}), i.e. when $C_r=0$. This clearly demonstrates the importance of using $C_r$ over $C_s$.
	\item When the feature space used is $\mathcal{F}_2$, Co-training (SVM) is the best method. Next best methods are supervised learners KNN and Gaussian NB. Three remarks can be made here. First, our approach could not always defeat supervised learners as it is seen from tables~\ref{tb:auc_acc} and~\ref{tb:prf_F2}. This is not surprising and in fact lies at the inherent difference between semi-supervised and supervised methods-- unlabeled examples could make the trained model susceptible to error propagation and thus wrong estimation. Second, as it is seen in tables~\ref{tb:prf_F1}-\ref{tb:prf_F1_F2}, achieving very high recall on the negative examples and low score on the positive ones shall not be treated as a potent property, otherwise a trivial classifier which always assigns negative labels to all samples would be the best learner. Third, using $C_r$ always improves the performance over $C_s$. One point that needs to be clarified is, our ultimate goal is not to achieve high performance on the labeled data, but rather to detect the suspicious (unlabeled) advertisements which could be human trafficking related-- this will be explained in more details in~\ref{sec:blind}.   	
	\item Compared to the other semi-supervised approaches, our approach either achieved higher or comparable AUC scores. The reason we performed exactly the same as the Laplacian SVM, is because by setting $C_r=0$, the two approaches are inherently the same. 	
	\item For the Laplacian SVM to be able to run on $\mathcal{F}_1$, the Laplacian $\mathcal{L}'$ has to be constructed using $\mathcal{F}_1$ while inherently is supposed to be made using $\mathcal{F}_2$. This is because $C_r$ is essentially associated with $\mathcal{F}_1$, and $C_s$ corresponds to $\mathcal{L}'$ and correspondingly $\mathcal{F}_2$. The same holds for $\{\mathcal{F}_1,\mathcal{F}_2\}$, where we need to construct a new feature space by concatenating $\mathcal{F}_1$ and $\mathcal{F}_2$ as the Laplacian SVM does not inherently use $\mathcal{F}_1$ at all. The new feature space is then used to construct the Laplacian $\mathcal{L}'$.	
	\item Since our approach inherently incorporates both of the Laplacian matrices corresponding to the two feature spaces $\mathcal{F}_1$ and $\mathcal{F}_2$, all other baselines were also run using the concatenation of these two feature spaces for the sake of fair comparison. Unlike our approach which used the wise combination of $\mathcal{F}_1$ and $\mathcal{F}_2$, other methods do not gain high AUC by simply combining the feature spaces.
\end{itemize}

\begin{table}[t]
	\centering
	\caption{AUC and accuracy results with 10-fold cross-validation on the labeled data. The best performance is in bold.}
	\begin{tabular}{|c|c|c|c|c|c|c|c|} 
		\cline{1-8}
		\textbf{Learner} & \multicolumn{3}{c|}{\textbf{AUC}} & \multicolumn{3}{c|}{\textbf{Accuracy}}\\
		\hhline{========}
		& $\mathcal{F}_1$ & $\mathcal{F}_2$ & $\{\mathcal{F}_1, \mathcal{F}_2\}$ & $\mathcal{F}_1$ & $\mathcal{F}_2$ & $\{\mathcal{F}_1, \mathcal{F}_2\}$ \\ \cline{2-8}
		$S^3VM-R$ &  \textbf{0.91} & 0.9 & \textbf{0.96}  & \textbf{0.91} & 0.9  & \textbf{0.97}\\ \hline
		Laplacian SVM &  0.9 & 0.9 & 0.9  & 0.91 & 0.9 & 0.92\\ \hline
		LabelSpreading (RBF) & 0.78 & 0.87 & 0.84 & 0.8 & 0.85 & 0.86 \\ \hline
		LabelSpreading (KNN) & 0.68 & 0.80  & 0.74 & 0.71 & 0.8 & 0.8 \\ \hline
		Co-Training (SVM) & 0.82 & \textbf{0.94}  & 0.92 & 0.85 & \textbf{0.94} & 0.93 \\ \hline
		SVM & 0.82 & 0.9 & 0.91 & 0.85 & 0.92 & 0.93 \\ \hline
		KNN & 0.76 & 0.91 & 0.81 & 0.79 & 0.92 & 0.84 \\ \hline
		Gaussian NB & 0.78 & 0.91 & 0.9 & 0.82 & 0.9 & 0.9 \\ \hline
		Logistic Regression & 0.82 & 0.89 & 0.88 & 0.85 & 0.92 & 0.92 \\ \hline
		AdaBoost & 0.82 & 0.85 & 0.85 & 0.85 & 0.88 & 0.88 \\ \hline
		Random Forest & 0.81 & 0.89  & 0.89 & 0.83 & 0.91 & 0.92\\ \hline
	\end{tabular}
	\label{tb:auc_acc}
\end{table}

\begin{table}[t]
	\centering
	\caption{Precision, recall and F1-score for the positive and negative classes using $\mathcal{F}_1$. Experiments were run using 10-fold cross-validation on the labeled data. The best performance is in bold.}
	\begin{tabular}{|c|c|c|c|c|c|c|} 
		\cline{1-7}
		\textbf{Learner} & \multicolumn{2}{c|}{\textbf{Precision}} &
		\multicolumn{2}{c|}{\textbf{Recall}} &
		\multicolumn{2}{c|}{\textbf{F1-score}}\\
		\hhline{=======}
		& $class_p$ & $class_n$ & $class_p$ & $class_n$ & $class_p$ & $class_n$\\ \cline{2-7}
		$S^3VM-R$ & \textbf{0.91} & \textbf{0.92} & \textbf{0.91} & \textbf{0.93} & \textbf{0.91} & \textbf{0.92}\\ \hline
		Laplacian SVM & 0.86 & 0.89 & 0.88 & 0.9 & 0.87 & 0.88 \\ \hline
		LabelSpreading (RBF) & 0.76 & 0.78 & 0.77 & 0.73 & 0.8 & 0.81\\ \hline
		LabelSpreading (KNN) & 0.65 & 0.7 & 0.71 & 0.68 & 0.69 & 0.73 \\ \hline
		Co-Training (SVM) & 0.81 & 0.84 & 0.71 & 0.92 & 0.73 & 0.87  \\ \hline
		SVM & 0.86 & 0.83 & 0.68 & 0.96 & 0.74 & 0.88 \\ \hline
		KNN & 0.72 & 0.8 & 0.63 & 0.88 & 0.65 & 0.83 \\ \hline
		Gaussian NB & 0.79 & 0.81 & 0.72 & 0.85 & 0.73 & 0.81 \\ \hline
		Logistic Regression & 0.81 & 0.85 & 0.71 & 0.93 & 0.74 & 0.88\\ \hline
		AdaBoost & 0.86 & 0.83 & 0.68 & 0.95 & 0.74 & 0.88\\ \hline
		Random Forest & 0.77 & 0.85 & 0.73 & 0.89 & 0.73 & 0.86\\ \hline
	\end{tabular}
	\label{tb:prf_F1}
\end{table}

\begin{table}[t]
	\centering
	\caption{Precision, recall and F1-score for the positive and negative classes using $\mathcal{F}_2$. Experiments were run using 10-fold cross-validation on the labeled data. The best performance is in bold.}
	\begin{tabular}{|c|c|c|c|c|c|c|} 
		\cline{1-7}
		\textbf{Learner} & \multicolumn{2}{c|}{\textbf{Precision}} &
		\multicolumn{2}{c|}{\textbf{Recall}} &
		\multicolumn{2}{c|}{\textbf{F1-score}}\\
		\hhline{=======}
		& $class_p$ & $class_n$ & $class_p$ & $class_n$ & $class_p$ & $class_n$\\ \cline{2-7}
		$S^3VM-R$ & 0.91 & 0.9 & 0.9 & 0.9 & 0.91 & 0.92 \\ \hline
		Laplacian SVM & 0.91 & 0.9 & 0.9 & 0.91 & 0.89 & 0.92\\ \hline
		LabelSpreading (RBF) & 0.8 & 0.86 & 0.82 & 0.83 & 0.81 & 0.85 \\ \hline
		LabelSpreading (KNN) & 0.7 & 0.75 & 0.73 & 0.78 & 0.79 & 0.77 \\ \hline
		Co-Training (SVM) & \textbf{0.96} & 0.91 &  0.91 & \textbf{0.97} & 0.93 & \textbf{0.93} \\ \hline
		SVM & 0.93 & 0.91 & 0.84 & \textbf{0.97} & 0.87 & \textbf{0.93} \\ \hline
		KNN & 0.87 & 0.92 & 0.88 & 0.94 & 0.87 & \textbf{0.93} \\ \hline
		Gaussian NB & 0.78 & \textbf{0.96} & \textbf{0.94} & 0.87 & 0.84 & 0.91 \\ \hline
		Logistic Regression & 0.98 & 0.89 & 0.81 & 0.98 & 0.88 & \textbf{0.93} \\ \hline
		AdaBoost & 0.88 & 0.88 & 0.75 & 0.95 & 0.78 & 0.91\\ \hline
		Random Forest & 0.93 & 0.89 & 0.81 & \textbf{0.97} & 0.85 & \textbf{0.93}\\ \hline
	\end{tabular}
	\label{tb:prf_F2}
\end{table}

\begin{table}[t]
	\centering
	\caption{Precision, recall and F1-score for the positive and negative classes using $\{\mathcal{F}_1, \mathcal{F}_2\}$. Experiments were run using 10-fold cross-validation on the labeled data. The best performance is in bold.}
	\begin{tabular}{|c|c|c|c|c|c|c|} 
		\cline{1-7}
		\textbf{Learner} & \multicolumn{2}{c|}{\textbf{Precision}} &
		\multicolumn{2}{c|}{\textbf{Recall}} &
		\multicolumn{2}{c|}{\textbf{F1-score}}\\
		\hhline{=======}
		& $class_p$ & $class_n$ & $class_p$ & $class_n$ & $class_p$ & $class_n$\\ \cline{2-7}
		$S^3VM-R$ & \textbf{0.97} & \textbf{0.97} & \textbf{0.95} & \textbf{0.98} & \textbf{0.94} & \textbf{0.95}\\ \hline
		Laplacian SVM & 0.96 & 0.94 & 0.91 & 0.96 & 0.91 & 0.93 \\ \hline
		LabelSpreading (RBF) & 0.83 & 0.86 & 0.82 & 0.84 & 0.81 & 0.86\\ \hline
		LabelSpreading (KNN) & 0.71 & 0.74 & 0.75 & 0.78 & 0.8 & 0.78\\ \hline
		Co-Training (SVM) & 0.92 & 0.9 & 0.9 & 0.94 & 0.91 & 0.92  \\ \hline
		SVM & 0.96 & 0.92 & 0.84 & 0.97 & 0.89 & 0.94 \\ \hline
		KNN & 0.84 & 0.83 &0.67 & 0.95 & 0.73 & 0.88 \\ \hline
		Gaussian NB & 0.77 & 0.96 & 0.94 & 0.87 & 0.84 & 0.91 \\ \hline
		Logistic Regression & 0.95 & 0.9 & 0.79 & 0.97 & 0.85 & 0.93 \\ \hline
		AdaBoost & 0.88 & 0.88 & 0.75 & 0.95 & 0.78 & 0.91\\ \hline
		Random Forest & 0.93 & 0.9 & 0.82 & 0.97 & 0.86 & 0.93 \\ \hline
	\end{tabular}
	\label{tb:prf_F1_F2}
\end{table}

\subsubsection{Blind Evaluation}\label{sec:blind}
For the next set of experiments, we first run our method on the \textit{entire} filtered dataset and without cross-validation. Recall from the previous sections that this is to make a better use of the unlabeled examples. Then the following \textit{control} experiment was conducted. Our learner was tested on the whole set of the unlabeled examples. Out of 3,343 instances, our approach identified two sets of positive and negative instances. The positive set contained 394 advertisements which were likely to be of interest to law enforcement, whereas the negative set included the remaining 2,962 unlabeled advertisements of probably less interest to law enforcement. Next, to precisely determine the correctly identified fractions of these two sets, we randomly picked two subsets (control groups) of 100 examples from each set for further validation by our expert.

We passed these two control groups to our expert for further verification. The expert-validated results demonstrated that all of the examples in the positive group were of interest to law enforcement, while only two examples from the negative group were not correctly classified as of not being of any interest to law enforcement. Thus, both results support the effectiveness of our framework in identifying highly human trafficking advertisements. Using the same two control groups and AUC metric, we now perform so-called blind evaluation (see table~\ref{tb:blind}) of other baselines. Note, we call this blind since actual labels are not provided and the expert-generated labels might convey uninformative information. In general, supervised methods failed to achieve good results in the blind evaluation compared to most of the semi-supervised methods.

\begin{table}[!h]
	\centering
	\caption{Blind evaluation of the baselines on the two control groups. The best performance is in bold.}
	\begin{tabular}{|c|c|c|c|c|} 
		\cline{1-5}
		\textbf{Learner} & \multicolumn{3}{c|}{\textbf{AUC}} \\
		\hhline{=====}
		& $\mathcal{F}_1$ & $\mathcal{F}_2$ & $\{\mathcal{F}_1, \mathcal{F}_2\}$ \\ \cline{2-5}
		Laplacian SVM &  \textbf{0.9} & \textbf{0.92} & \textbf{0.93}\\ \hline
		LabelSpreading (RBF) & 0.75 & 0.85 & 0.87\\ \hline
		LabelSpreading (KNN) & 0.7 & 0.82  & 0.79\\ \hline
		Co-Training (SVM) & 0.8 & 0.9  & 0.91\\ \hline
		SVM & 0.8 & 0.65 & 0.69\\ \hline
		KNN & 0.74 & 0.62 & 0.77\\ \hline
		Gaussian NB & 0.77 & 0.51 & 0.52\\ \hline
		Logistic Regression & 0.76 & 0.62 & 0.75\\ \hline
		AdaBoost & 0.77 & 0.74 & 0.74\\ \hline
		Random Forest & 0.8 & 0.8  & 0.8\\ \hline
	\end{tabular}
	\label{tb:blind}
\end{table} 

\subsection{Hyperparameter Sensitivity}
\begin{figure*}[!ht]
	\caption{Effect of varying different parameters on the performance.}
	\centering
	\includegraphics[width=0.3\textwidth]{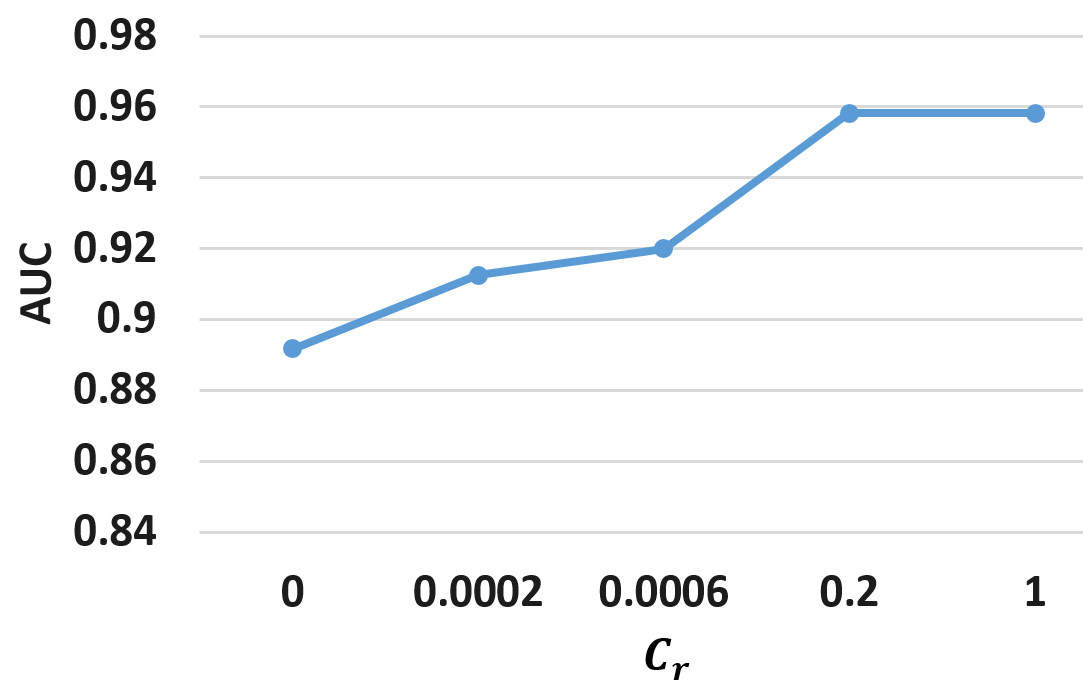}
	\includegraphics[width=0.3\textwidth]{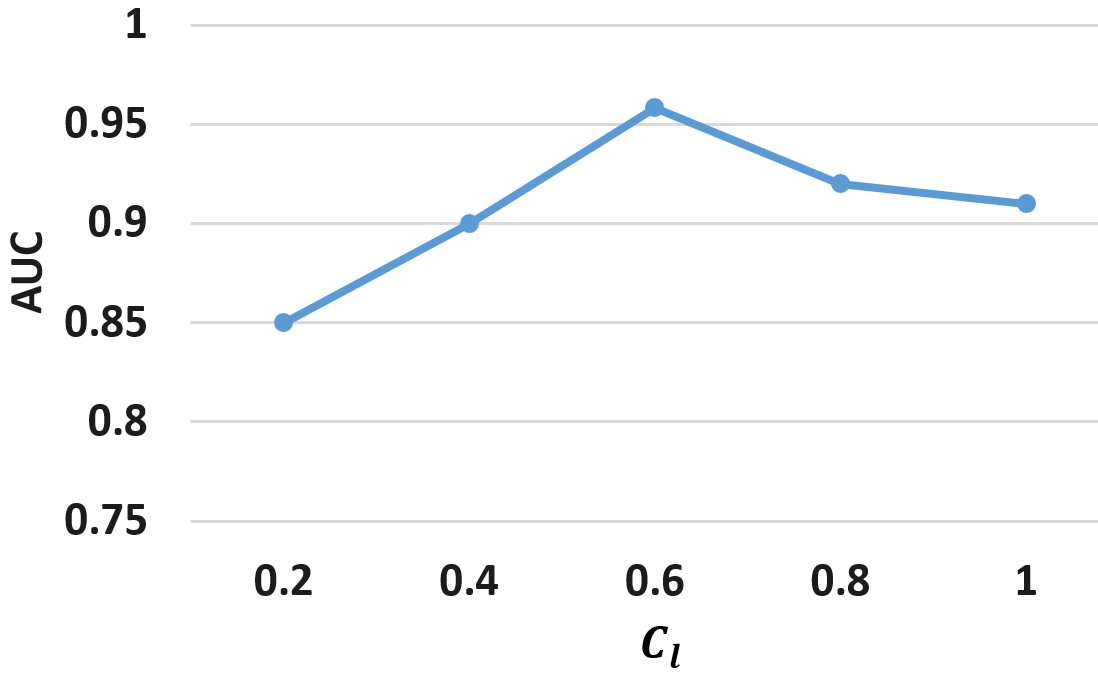}
	\includegraphics[width=0.3\textwidth]{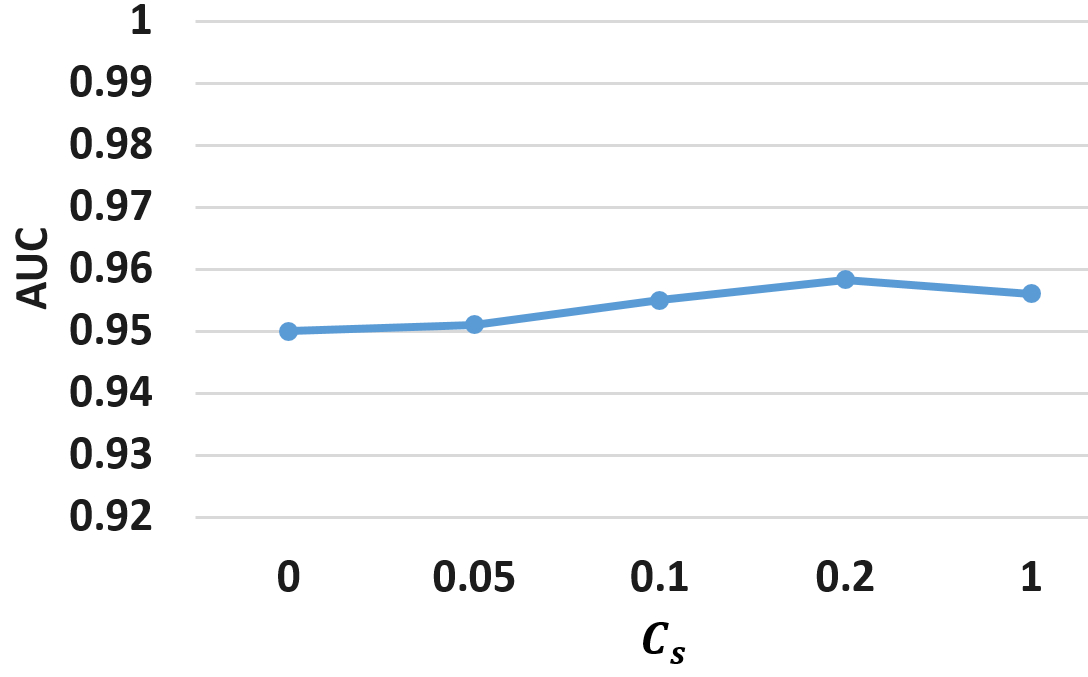}	
	\label{fig:para_sens}
\end{figure*}

Here, we discuss how altering the hyperparameters $C_l$, $C_r$ and $C_s$ may affect the performance of $S^3VM-R$. We start off by fixing the value of $C_l$ to 0.6, which was empirically found to work well in our experiments. Also, recall from the previous sections that one typical choice for $C_s$ is $\frac{\gamma_I}{(l+u)^2}$~\cite{belkin2006manifold}. Here, we set $C_s=0.2$ and varied the values of $C_r$ as $\{0, 0.0002, 0.0006, 0.2, 1.0\}$ and plotted the results in Figure~\ref{fig:para_sens}. We used the same 10-fold cross-validation setting from the previous section. 

We made the following observation. With the slight increase of $C_r$, the performance of our approach increased, peaked and then stabilized, i.e., further increase of $C_r$ did not change the performance. This suggests significance of deploying the additional information from our first feature space $\mathcal{F}_1$, over $\mathcal{F}_2$ and its corresponding smoothness penalty parameter $C_s$ which is used by $S^3VM-R$ and the standard Laplacian SVM. 

Next, to see the impact of $C_l$ on the performance, we set $C_r=0.2$ and varied $C_l$ as $\{0.2, 0.4, 0.6, 0.8, 1.0\}$. The results are depicted in Figure~\ref{fig:para_sens}. We note that setting $C_l=0$ is meaningless and thus we do not have any performance corresponding to that-- otherwise each $\beta_i$ in Eq.~\ref{eq:beta_star} would be zero. In general, the performance was not particular sensitive to this parameter-- varying by 0.2 for values of 0.4 and greater.
 
Finally, having fixed $C_l=0.6$ and $C_r=0.2$, we also tried other values for $C_s$ including $\sum_{i,j=1}^{l+u}W_{ij}$ suggested by~\cite{belkin2006manifold} and depicted the results in Figure~\ref{fig:para_sens}. The results suggest that our approach is less sensitive to this parameter compared to $C_r$ and $C_l$.


\label{sec:hypr}

\subsection{Significance of Features}
To examine how much discriminative our feature groups in $\mathcal{F}_1$ are, we further conducted an analysis using the labeled examples and the standard feature selection measure $\chi^2$ to find the top features-- only half of the features with scores greater than a given threshold (0.5) were selected (see table~\ref{tb:chi2} for the complete set of features and their corresponding $\chi^2$ scores).


From this list, we noticed that `countries of interest' and `reference to spa massage therapy' were the most discriminative feature groups, while `advertisement language pattern' group (with 3 important features) appeared to be the most dominant feature group. 

\begin{figure}[!ht]
	\caption{Frequency of each feature in $\mathcal{F}_1$ in the filtered dataset. Features are grouped into the two groups, most important ($\mathcal{F}^*_1$) and less important features ($\overline{\mathcal{F}^*_1}$), according to $\chi^2$.}
	\centering
	\includegraphics[width=0.7\textwidth]{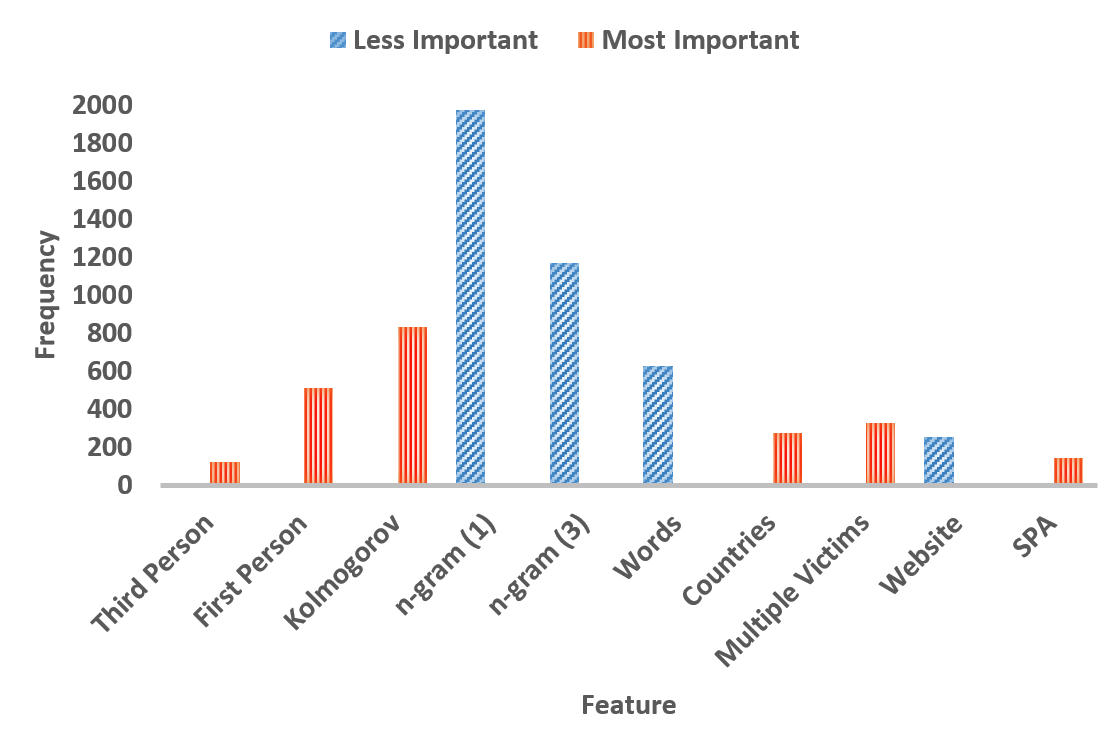}
	\label{fig:feat}
\end{figure}

Figure~\ref{fig:feat} compares the top features against the less important subset of the features (denoted by $\overline{\mathcal{F}^*_1}$) in the filtered dataset, in terms of frequency values. Note for clarity, we have removed from this figure, the features with frequency less than 20. According to this figure, our most discriminative features are not necessarily those that appear more often.

\begin{table}[t]
	\centering
	\caption{Significance of the features in $\mathcal{F}_1$. The check-marked features show the top features.}
	\begin{tabular}{|c|l|c|c|}
		\hline
		\textbf{No.} & \textbf{Feature Group} & $\chi^2$ & \textbf{Selected}\\ \hline\hline
		\multirow{5}{*}{1} & \textbf{Advertisement Language Pattern} & &\\
		& \tab - Third person language & 8.4 & \checkmark\\
		& \tab - First person plural pronouns & 9.5 & \checkmark \\
		& \tab - Kolmogorov complexity & 0.7 & \checkmark\\ 
		& \tab - \textit{n}-grams (1) & 0.4 &\\
		& \tab - \textit{n}-grams (2) & 0.0 &\\
		& \tab - \textit{n}-grams (3) & 0.4 &\\ 
		& & &\\
		2 & \textbf{Words and Phrases of Interest} & 0.0 &\\
		& & &\\
		3 & \textbf{Countries of Interest} & 59.3 & \checkmark \\
		& & &\\
		4 & \textbf{Multiple Victims Advertised} & 14.1 &\checkmark\\
		& & &\\
		5 & \textbf{Victim Weight} & 0.2 &\\
		& & &\\
		\multirow{2}{*}{6} & \textbf{Reference to Website or Spa Massage Therapy} & &\\
		& \tab - Reference to website & 0.1 &\\
		& \tab - Reference to Spa Massage Therapy & 33.5 & \checkmark\\ \hline
	\end{tabular}
	\label{tb:chi2}
\end{table}

To further investigate the importance of each of the top features, we performed classification using the labeled examples and the previous setting, on basis of these two subsets of the features and their combination, i.e., $\mathcal{F}^*_1$, $\overline{\mathcal{F}^*_1}$ and $\mathcal{F}_1$. The classification results are shown in Table~\ref{tb:diff_feat}. We made the following observations: 
\begin{itemize}
	\item Considering only the feature space $\mathcal{F}_1$, our approach achieved higher performance compared to all other baselines by either using the whole feature space or the most discriminative features $\mathcal{F}^*_1$.
	\item Deploying only the features from $\mathcal{F}^*_1$, we were able to achieve comparable results as if we used the whole feature space $\mathcal{F}_1$.   
\end{itemize}

\begin{table}[h!]
	\centering
	\caption{Classification results (AUC) using 10-fold cross-validation and different subsets of the features on the labeled data.}
	\begin{tabular}{|l|c|c|c|}
		\cline{1-4}
		\textbf{Name}          & \multicolumn{3}{c|}{\textbf{Value}}\\
		\hhline{====}
		  & $\overline{\mathcal{F}^*_1}$ & $\mathcal{F}^*_1$ & $\mathcal{F}_1$ \\ \cline{2-4}
		$S^3VM-R$  & 0.82 & 0.87 & 0.91 \\
		\cline{1-4}
	\end{tabular}
	\label{tb:diff_feat}
\end{table}

\section{Conclusion}
Readily available online data from escort advertisements could be leveraged in favor of fight against human trafficking. In this study, having focused on textual information from the available data crawled from Backpage.com, we identified if an escort advertisement can be reflective of human trafficking activities. In particular, we first proposed an unsupervised filtering approach to filter out the data which are more likely involved in human trafficking. We then proposed a semi-supervised learner, namely $S^3VM-R$, and trained it on a small portion of the data which was hand-labeled by a human trafficking expert. We used the trained model to identify labels of unseen data. Results suggested our approach is effective at identifying potential human trafficking related advertisements.

Our future plans include replicating the study by integrating more interesting features especially those supported by the criminology literature. Also, since hand-labeling unlabeled examples is expensive, an interesting research direction would be to deploy active learning to enable iterative supervised learning to actively query the user for labels. We also note that real-world data is often more imbalanced compared to our data, and the reason is that number of negative samples usually outweigh positive ones. We would thus like to apply the proposed framework on a more realistic dataset which contains much less suspicious posts than normal posts.

\section{Competing Interests}
The authors declare that they have no competing interests.

\section{Authors' Contributions}
HA developed and implemented the human trafficking detection approach and drafted the manuscript. PS provided guidance through the whole project and revised the manuscript. JS was in contact with an expert from law enforcement who was responsible for hand-labeling portions of the data. All authors read and approved the final manuscript.

\section*{Acknowledgment}
This work was funded by the Find Me Group, a 501(c)3 dedicated to bring resolution and closure to families of missing persons. The authors would like to thank anonymous reviewers for their valuable suggestions to improve the quality of the paper.



%
%

\bibliographystyle{IEEEtran}
\bibliography{IEEEfull,ref}

\begin{thebibliography}{10}
\providecommand{\url}[1]{#1}
\csname url@samestyle\endcsname
\providecommand{\newblock}{\relax}
\providecommand{\bibinfo}[2]{#2}
\providecommand{\BIBentrySTDinterwordspacing}{\spaceskip=0pt\relax}
\providecommand{\BIBentryALTinterwordstretchfactor}{4}
\providecommand{\BIBentryALTinterwordspacing}{\spaceskip=\fontdimen2\font plus
\BIBentryALTinterwordstretchfactor\fontdimen3\font minus
  \fontdimen4\font\relax}
\providecommand{\BIBforeignlanguage}[2]{{%
\expandafter\ifx\csname l@#1\endcsname\relax
\typeout{** WARNING: IEEEtran.bst: No hyphenation pattern has been}%
\typeout{** loaded for the language `#1'. Using the pattern for}%
\typeout{** the default language instead.}%
\else
\language=\csname l@#1\endcsname
\fi
#2}}
\providecommand{\BIBdecl}{\relax}
\BIBdecl

\bibitem{unodc}
\BIBentryALTinterwordspacing
``{UNODC on human trafficking and migrant smuggling},'' 2011. [Online].
  Available: \url{https://www.unodc.org/unodc/en/human-trafficking/}
\BIBentrySTDinterwordspacing

\bibitem{tvpa2000}
\BIBentryALTinterwordspacing
``{Trafficking victims protection act of 2000},'' 2000. [Online]. Available:
  \url{https://www.state.gov/j/tip/laws/61124.htm}
\BIBentrySTDinterwordspacing

\bibitem{HT_report2015}
\BIBentryALTinterwordspacing
``{Trafficking in persons report},'' July 2015. [Online]. Available:
  \url{https://www.state.gov/j/tip/rls/tiprpt/2015/}
\BIBentrySTDinterwordspacing

\bibitem{desplaces92}
C.~Desplaces, ``{Police run `Prostitution' sting; 19 men arrested, charged in
  Fourth East Dallas operation.}'' Nov Dallas Morning News, 21 Nov. 1992, 34A.
  Web. 24 Apr. 2012.

\bibitem{nicholas2012}
\BIBentryALTinterwordspacing
N.~D. Kristof, ``{How pimps use the web to sell girls.}'' Jan 2012. [Online].
  Available:
  \url{http://www.nytimes.com/2012/01/26/opinion/how-pimps-use-the-web-to-sell-girls.html}
\BIBentrySTDinterwordspacing

\bibitem{kennedy2012predictive}
E.~Kennedy, ``{Predictive patterns of sex trafficking online},'' Dietrich
  College Honors Theses, 2012.

\bibitem{mitchell2006learning}
T.~M. Mitchell, ``Learning from labeled and unlabeled data,'' \emph{Machine
  learning}, vol.~10, p. 701, 2006.

\bibitem{beigi2016signed}
G.~Beigi, J.~Tang, and H.~Liu, ``Signed link analysis in social media
  networks,'' in \emph{International AAAI Conference on Web and Social Media
  (ICWSM)}, 2016.

\bibitem{backstrom2011supervised}
L.~Backstrom and J.~Leskovec, ``Supervised random walks: predicting and
  recommending links in social networks,'' in \emph{Proceedings of the fourth
  ACM international conference on Web search and data mining}.\hskip 1em plus
  0.5em minus 0.4em\relax ACM, 2011, pp. 635--644.

\bibitem{alvari2016identifying}
\BIBentryALTinterwordspacing
H.~Alvari, A.~Hajibagheri, G.~Sukthankar, and K.~Lakkaraju, ``Identifying
  community structures in dynamic networks,'' \emph{Social Network Analysis and
  Mining}, vol.~6, no.~1, p.~77, 2016. [Online]. Available:
  \url{http://dx.doi.org/10.1007/s13278-016-0390-5}
\BIBentrySTDinterwordspacing

\bibitem{beigi2016exploiting}
G.~Beigi, J.~Tang, S.~Wang, and H.~Liu, ``Exploiting emotional information for
  trust/distrust prediction,'' in \emph{Proceedings of the 2016 SIAM
  International Conference on Data Mining (ICDM)}, 2016.

\bibitem{mitchell1997machine}
T.~Mitchell, \emph{Machine Learning}.\hskip 1em plus 0.5em minus 0.4em\relax
  McGraw-Hill Education, 1997.

\bibitem{beigi2014leveraging}
G.~Beigi, M.~Jalili, H.~Alvari, and G.~Sukthankar, ``Leveraging community
  detection for accurate trust prediction,'' in \emph{In ASE International
  Conference on Social Computing, Palo Alto, CA}, May 2014.

\bibitem{belkin2006manifold}
M.~Belkin, P.~Niyogi, and V.~Sindhwani, ``Manifold regularization: A geometric
  framework for learning from labeled and unlabeled examples,'' \emph{Journal
  of machine learning research}, vol.~7, no. Nov, pp. 2399--2434, 2006.

\bibitem{alvari2016non}
H.~Alvari, P.~Shakarian, and J.~Snyder, ``A non-parametric learning approach to
  identify online human trafficking,'' \emph{Intelligence and Security
  Informatics (ISI), 2016 IEEE Conference on}, pp. 133–--138, 2016.

\bibitem{hughes2005demand}
D.~M. Hughes \emph{et~al.}, ``{The demand for victims of sex trafficking},''
  \emph{Women’s Studies Program, University of Rhode Island}, 2005.

\bibitem{hughes2002use}
D.~M. Hughes, ``{The use of new communications and information technologies for
  sexual exploitation of women and children},'' \emph{Hastings Women's Law
  Journal}, vol.~13, no.~1, pp. 129--148, 2002.

\bibitem{latonero2011human}
M.~Latonero, ``{Human trafficking online: The role of social networking sites
  and online classifieds},'' \emph{Available at SSRN 2045851}, 2011.

\bibitem{dominique2015}
\BIBentryALTinterwordspacing
D.~{Roe-­‐Sepowitz}, J.~{Gallagher}, K.~{Bracy}, L.~{Cantelme},
  A.~{Bayless}, J.~{Larkin}, A.~{Reese}, and L.~{Allbee}, ``{Exploring the
  impact of the super bowl on sex trafficking},'' Feb. 2015. [Online].
  Available:
  \url{https://www.mccaininstitute.org/exploring-the-impact-of-the-super-bowl-on-sex-trafficking-2015/}
\BIBentrySTDinterwordspacing

\bibitem{2016arXiv160205048M}
\BIBentryALTinterwordspacing
K.~{Miller}, E.~{Kennedy}, and A.~{Dubrawski}, ``{Do public events affect sex
  trafficking activity?}'' \emph{ArXiv e-prints}, Feb. 2016. [Online].
  Available: \url{https://arxiv.org/abs/1602.05048}
\BIBentrySTDinterwordspacing

\bibitem{conf/semweb/SzekelyKSPSYKNM15}
P.~A. Szekely, C.~A. Knoblock, J.~Slepicka, A.~Philpot, A.~Singh, C.~Yin,
  D.~Kapoor, P.~Natarajan, D.~Marcu, K.~Knight, D.~Stallard, S.~S.
  Karunamoorthy, R.~Bojanapalli, S.~Minton, B.~Amanatullah, T.~Hughes,
  M.~Tamayo, D.~Flynt, R.~Artiss, S.-F. Chang, T.~Chen, G.~Hiebel, and
  L.~Ferreira, ``{Building and using a knowledge graph to combat human
  trafficking.}'' in \emph{International Semantic Web Conference (2)}, ser.
  Lecture Notes in Computer Science, vol. 9367.\hskip 1em plus 0.5em minus
  0.4em\relax Springer, 2015, pp. 205--221.

\bibitem{2015arXiv150906659N}
\BIBentryALTinterwordspacing
C.~{Nagpal}, K.~{Miller}, B.~{Boecking}, and A.~{Dubrawski}, ``{An entity
  resolution approach to isolate instances of human trafficking online},''
  \emph{ArXiv e-prints}, Sep. 2015. [Online]. Available:
  \url{https://arxiv.org/abs/1509.06659}
\BIBentrySTDinterwordspacing

\bibitem{doi:10.1080/23322705.2015.1015342}
A.~Dubrawski, K.~Miller, M.~Barnes, B.~Boecking, and E.~Kennedy, ``Leveraging
  publicly available data to discern patterns of human-trafficking activity,''
  \emph{Journal of Human Trafficking}, vol.~1, no.~1, pp. 65--85, 2015.

\bibitem{Zhou04learningwith}
D.~Zhou, O.~Bousquet, T.~N. Lal, J.~Weston, and B.~Schölkopf, ``Learning with
  local and global consistency,'' in \emph{Advances in Neural Information
  Processing Systems 16}.\hskip 1em plus 0.5em minus 0.4em\relax MIT Press,
  2004, pp. 321--328.

\bibitem{Li:2008:IKC:1478784}
M.~Li and P.~M. Vit\'{a}nyi, \emph{{An introduction to kolmogorov complexity
  and its applications}}, 3rd~ed.\hskip 1em plus 0.5em minus 0.4em\relax
  Springer Publishing Company, Incorporated, 2008.

\bibitem{conf/setn/KanarisKS06}
I.~Kanaris, K.~Kanaris, and E.~Stamatatos, ``{Spam detection using character
  n-grams.}'' in \emph{SETN}, ser. Lecture Notes in Computer Science, vol.
  3955.\hskip 1em plus 0.5em minus 0.4em\relax Springer, 2006, pp. 95--104.

\bibitem{hetter2012}
\BIBentryALTinterwordspacing
K.~Hetter, ``{Fighting sex trafficking in hotels, one room at a time},'' March
  2012. [Online]. Available:
  \url{http://www.cnn.com/2012/02/29/travel/hotel-sex-trafficking/}
\BIBentrySTDinterwordspacing

\bibitem{Lloyd2012}
\BIBentryALTinterwordspacing
R.~Lloyd, ``{An open letter to jim Buckmaster},'' April 2012. [Online].
  Available:
  \url{http://www.huffingtonpost.com/rachel-lloyd/an-open-letter-to-jim-buc\_b\_570666.html}
\BIBentrySTDinterwordspacing

\bibitem{Goodman2011}
\BIBentryALTinterwordspacing
J.~{Dickinson Goodman} and M.~{Holmes}, ``{Can we use RSS to catch rapists},''
  2011. [Online]. Available:
  \url{http://jessicadickinsongoodman.com/2011/10/30/can-we-use-rss-to-catch-rapists-poster-finished/}
\BIBentrySTDinterwordspacing

\bibitem{weightchart}
\BIBentryALTinterwordspacing
``{Average height to weight chart - babies to teenagers}.'' [Online].
  Available:
  \url{http://www.disabled-world.com/artman/publish/height-weight-teens.shtml}
\BIBentrySTDinterwordspacing

\bibitem{shannon2001mathematical}
C.~E. Shannon, ``A mathematical theory of communication,'' \emph{ACM SIGMOBILE
  Mobile Computing and Communications Review}, vol.~5, no.~1, pp. 3--55, 2001.

\bibitem{maaten2008visualizing}
L.~van~der Maaten and G.~Hinton, ``{Visualizing high-dimensional data using
  t-SNE},'' \emph{Journal of Machine Learning Research, vol. 9}, pp.
  2579--2605, 2008.

\bibitem{cortes1995support}
C.~Cortes and V.~Vapnik, ``Support-vector networks,'' \emph{Machine learning},
  vol.~20, no.~3, pp. 273--297, 1995.

\bibitem{belkin2005manifold}
M.~Belkin, P.~Niyogi, and V.~Sindhwani, ``On manifold regularization.'' in
  \emph{Proceedings of the Tenth International Workshop on Artificial
  Intelligence and Statistics (AISTAT)}, 2005.

\bibitem{grigor2006heat}
A.~Grigor’yan, ``Heat kernels on weighted manifolds and applications,''
  \emph{Cont. Math}, vol. 398, pp. 93--191, 2006.

\bibitem{blum1998combining}
A.~Blum and T.~Mitchell, ``Combining labeled and unlabeled data with
  co-training,'' in \emph{Proceedings of the eleventh annual conference on
  Computational learning theory}.\hskip 1em plus 0.5em minus 0.4em\relax ACM,
  1998, pp. 92--100.

\end{thebibliography}

\end{document}